\documentclass[letterpaper, 10 pt, conference]{ieeeconf}  

\IEEEoverridecommandlockouts                              
\usepackage[l2tabu,orthodox]{nag}

\usepackage[
    backend=bibtex8,
    style=ieee,
    sorting=none,
    natbib=true,
    doi=false,
    isbn=false,
    url=false,
    eprint=false,
    maxcitenames=1,
    mincitenames=1
]{biblatex}

\usepackage[pdftex,colorlinks]{hyperref}

\usepackage[printonlyused]{acronym}

\usepackage{siunitx}
\usepackage[all]{nowidow}

\usepackage[dvipsnames]{xcolor}

\usepackage{lipsum}

\usepackage{xspace} 

\usepackage[pdftex]{graphicx}

\usepackage{epstopdf}

\usepackage{import}

\graphicspath{{./latexGoodPractices/}}

\usepackage{booktabs}

\usepackage{tabularx}
\usepackage{multirow, multicol}

\usepackage{amssymb,amsfonts,amsmath,amscd}

\usepackage{bm}

\usepackage{soul}

\newcommand{\bbm}{\begin{bmatrix}}
\newcommand{\ebm}{\end{bmatrix}}

\title{\LARGE \bf 
    DRIVE: Data-driven Robot Input Vector Exploration
}

\author{Dominic Baril$^{1}$, Simon-Pierre Deschênes, Luc Coupal, Cyril Goffin, \\ Julien Lépine, Philippe Giguère, François Pomerleau$^{1}$
\thanks{*This research was supported by the Fonds de recherche du Québec – Nature et technologies (FRQNT) and by the Natural Sciences and Engineering Research Council of Canada (NSERC) through grant CRDPJ 527642-18 SNOW (Self-driving Navigation Optimized for Winter).}
	\thanks{$^{1}$Northern Robotics Laboratory, Université Laval, Quebec City, Quebec, Canada
		{\texttt{\small \{ dominic.baril, francois.pomerleau \}@norlab.ulaval.ca}}}%
}

\usepackage{tikz}
\newcommand\copyrighttext{%
	\footnotesize \textcopyright 2024 IEEE. Personal use of this material is permitted. Permission from IEEE must be obtained for all other uses, in any current or future media, including reprinting/republishing this material for advertising or promotional purposes, creating new collective works, for resale or redistribution to servers or lists, or reuse of any copyrighted component of this work in other works.}
\newcommand\copyrightnotice{%
	\begin{tikzpicture}[remember picture,overlay]
		\node[anchor=south,yshift=15pt] at (current page.south) {\parbox{\dimexpr\textwidth-\fboxsep-\fboxrule\relax}{\copyrighttext}};
	\end{tikzpicture}%
}

\usepackage[switch]{lineno}

\acrodef{SLAM}{simultaneous localization and mapping}
\acrodef{SOTA}{state-of-the-art}
\acrodef{SSMR}{skid-steering mobile robot}
\acrodef{AMR}{Ackermann mobile robot}
\acrodef{UGV}{uncrewed ground vehicle} 
\acrodef{IDD}{ideal differential-drive}
\acrodef{ICR}{instantaneous center or rotation}
\acrodef{RTK}{Realtime Kinematics}
\acrodef{GNSS}{Global Navigation Satellite System}
\acrodef{ROC}{radius of curvature}
\acrodef{IMU}{inertial measurement unit}
\acrodef{MPC}{model predictive control}
\acrodef{GP}{Gaussian processe}
\acrodef{BLR}{Bayesian linear regression}
\acrodef{IPEM}{integrated prediction error minimization}
\acrodef{MLP}{multilayer perceptron}
\acrodef{ICP}{iterative closest point}
\acrodef{MRMSE}{multi-step root mean squared error}
\acrodef{T-MRMSE}{translational multi-step root mean squared error}
\acrodef{R-MRMSE}{rotational multi-step root mean squared error}
\acrodef{M-Z-score}{multi-step Z-score}
\acrodef{DRIVE}{Data-driven Robot Input Vector Exploration}
\acrodef{VISTA}{Vehicle Input Space Training Assistant}

\newcommand{\DOUGHNUTCALIB}{\ac{DRIVE}\xspace}

\newcommand{\WINSIZE}{h}
\newcommand{\TRAINDATA}{$\mathcal{D}$\xspace}

\newcommand{\PREDSTATE}{\,^\mapf\!\hat{\bm q}}
\newcommand{\MEASSTATE}{\,^\mapf\bm q}
\newcommand{\MARMOTTE}{HD2\xspace}
\newcommand{\CMDBODYVEL}{^{\robotf}\bm{f}}
\newcommand{\SLIPBODYVEL}{^{\robotf}\bm{g}}
\newcommand{\OBSERVEDSLIP}{\bm{g}}
\newcommand{\INPUTVECTOR}{\bm{u}}
\newcommand{\STATEPROPMAT}{_{\robotf}^{\mapf}\bm{T}\left(^\mapf\theta_t\right)}
\newcommand{\INPUTSPACE}{\mathcal{J}}
\newcommand{\BODYVELSPACE}{\mathcal{B}}
\newcommand{\CENTRIFUGAL}{\psi}

\newcommand{\robotstate}{\bm{q}}
\newcommand{\mapf}{\mathcal{G}} 
\newcommand{\robotf}{\mathcal{R}} 

\newcommand{\blrweights}{\bm{\gamma}}
\newcommand{\blrinputs}{^{\robotf}\bm{x}}
\newcommand{\MRSME}{\epsilon}

\makeatletter
\def\blx@err@patch#1{}
\makeatother

\begin{document}

\maketitle
\copyrightnotice
\thispagestyle{empty}
\pagestyle{empty}

\begin{abstract}

An accurate motion model is a fundamental component of most autonomous navigation systems.
While much work has been done on improving model formulation, no standard protocol exists for gathering empirical data required to train models.
In this work, we address this issue by proposing~\DOUGHNUTCALIB, a protocol that enables characterizing~\acp{UGV} input limits and gathering empirical model training data.
We also propose a novel learned slip approach outperforming similar acceleration learning approaches. 
Our contributions are validated through an extensive experimental evaluation, cumulating over~\SI[detect-weight,mode=text]{7}{\kilo\meter} and~\SI[detect-weight=true,mode=text]{1.8}{\hour} of driving data over three distinct~\acp{UGV} and four terrain types. 
We show that our protocol offers increased predictive performance over common human-driven data-gathering protocols.
Furthermore, our protocol converges with~\SI[detect-weight,mode=text]{46}{\second} of training data, almost four times less than the shortest human dataset gathering protocol. 
We show that the operational limit for our model is reached in extreme slip conditions encountered on surfaced ice. 
~\DOUGHNUTCALIB is an efficient way of characterizing~\ac{UGV} motion in its operational conditions.
Our code and dataset are both available online at this link: \url{https://github.com/norlab-ulaval/DRIVE}.


\end{abstract}



\section{Introduction}
\label{sec:intro}
The ability to model the motion of \acfp{UGV} is fundamental to enabling
localization~\citep{Dumbgen2023}, path planning~\citep{Takemura2021} and path following~\citep{Brunke2022}. 
Poor vehicle-terrain characterization will lead to significant modeling errors, potentially causing system failure~\citep{Seegmiller2016}.
With limited available information and sensory measurements on vehicle and ground properties, generating a reliable~\ac{UGV} motion model remains challenging. 
For most models, training on empirical data is required to reduce modeling error~\citep{Seegmiller2013}. 
This task requires deploying a~\ac{UGV} in its operational environment and manually drive it for an extended period~\citep{Williams2018}.
Since energy consumption and deployment time are critical for various~\ac{UGV} applications, facilitating this task is of high importance.
Additionally, standardizing this process could help engineers to ensure that their systems comply with the ISO 34502:2022(E) standard on autonomous navigation.\footnote[2]{“ISO 34502:2022(E): Road vehicles — Test scenarios for automated driving systems — Scenario-based safety evaluation framework”, 2022}

\begin{figure}[t!]
	\centering
    \includegraphics[width=0.48\textwidth]{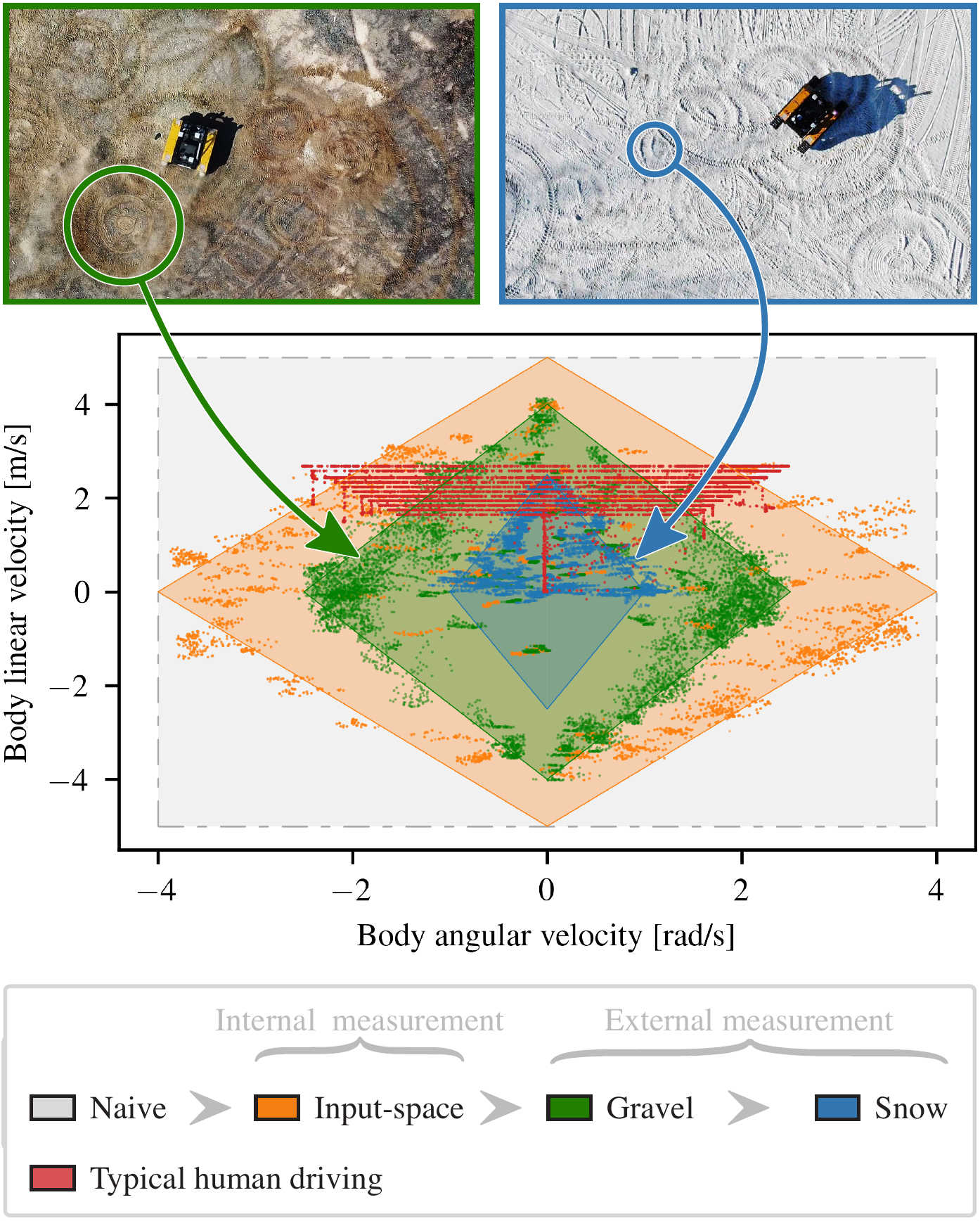}
	   \vspace{-0.1in}
    \caption{
            Vehicle and terrain characterization done through~\DOUGHNUTCALIB.
            The manufacturer-defined Naive input-space region is drawn in gray.
            The vehicle's true input-space, characterized through internal measurements, is shown in orange.
            Typical human driving is shown in red.
            The resulting body velocities are represented in green for gravel and blue for snow.
            }
	\label{fig:input-space}
\end{figure}

Most work on~\ac{UGV} motion modeling relies on manual driving to gather a training dataset, with little to no details on the driving protocol.
Thus, we propose the~\emph{\acf{DRIVE}}, a protocol aiming to facilitate and standardize vehicle characterization with respect to the terrain, as illustrated in~\autoref{fig:input-space}.
We start by identifying the true vehicle's input space, differing from the manufacturer's specifications.
We then automatically send commands to the~\ac{UGV} to cover the entire true input space.
This differs from the common manual driving approach, which tends to cover only forward driving, as shown by the red dots representing our previous work~\citep{Baril2020}.
We show that broad input-space coverage offers significant modeling error reduction compared to narrow coverage.
With this dataset, we train a learned vehicle slip model that maps~\ac{UGV} commands to resulting body velocities.
The resulting trained parameters vary significantly depending on terrain, as highlighted by the green and blue diamond areas in~\autoref{fig:input-space}, representing navigation on gravel and snow respectively.

The specific contributions of this paper are
(i)~\DOUGHNUTCALIB, a standardized~\ac{UGV} characterization and motion data generation protocol allowing motion models to be trained on the entire vehicle input space;
(ii) A novel slip-based \ac{UGV} motion prediction model, leveraging the accuracy of model-based approaches and the minimal system characterization requirement of learning-based approaches.
We validate our contributions with an extensive experimental evaluation featuring three distinct~\acp{UGV}, with weights ranging from~\SI{75}{\kilo\gram} to~\SI{470}{\kilo\gram}, two types of ground interaction (i.e., wheels and tracks) and four different terrain types.
Our observations rely on driving data totaling \SI{7}{\kilo\meter} and \SI{1.8}{\hour}.



\section{Related Work}
\label{sec:RW}
Most vehicle motion modeling approaches can be divided into two distinct categories: model-based and learning-based. 
Both categories share the requirement of using empirical driving data to train their parameters and reduce modeling errors.
For both categories, no standardized protocol exists for training dataset generation. 

\textbf{Model-based approaches} can be split into two distinct categories: \emph{kinematics} and \emph{dynamics}.
Kinematic models remain the most popular for~\acp{UGV} due to their low computational complexity and number of parameters to train.
For \acp{SSMR}, \citet{Mandow2007} reduced the model prediction error by~\SI{15}{\%} compared to the manufacturer's model using a kinematic model empirically identifying vehicle slip and skid.
\citet{Seegmiller2014} proposed a similar additive slip approach, computing slip based on kinematic quantities, yielding a prediction error reduction between~\SI{70}{\%} and~\SI{90}{\%} depending on terrain type, again compared to the manufacturer's model. 
\citet{Bussmann2018} extended the experimental validation for additive slip approaches and showed a similar performance for a~\SI{900}{\meter} experiment on off-road terrain. 
On the other hand, dynamic models account for various forces acting on the vehicle's body. 
\citet{Seegmiller2016} proposed a multi-body full dynamic motion model with a generic formulation based on vehicle geometry and properties.
This work has been extended by~\citet{Yang2022}, showing simulation errors of less than~\SI{3.4}{\%} for vehicle slip ratio. 
While being more accurate than kinematic models, dynamic models require extensive vehicle characterization effort and expertise.
For all of the work mentioned above, empirical training data is acquired through a human driving the~\ac{UGV} with little to no guidelines, which motivates our standardized protocol.

Alternatively, \textbf{learning-based approaches} have been explored in the literature, yielding more accurate models for extreme~\ac{UGV} motion.
In these approaches, part of the prediction is conducted through a nominal model, often represented by a unicycle model, with a module allowing system dynamics to be learned. 
\acp{GP} have become a popular approach to learn system dynamics, both for vehicle slip in off-road driving~\citep{Wang2023} and tire forces in high-speed road racing~\citep{Hewing2020}.
\citet{McKinnon2019} have proposed a similar approach, replacing \ac{GP} learning with \ac{BLR}. 
The lower computational complexity of \ac{BLR}, makes it a more suitable approach for real-time \ac{UGV} motion prediction.
Alternatively,~\citet{Djeumou2023} have proposed a tire-force learning framework allowing autonomous drifting to be performed with~\SI{3}{\minute} of driving data.
Deep learning has also been explored for motion prediction in off-road terrain. 
\citet{Williams2018} have shown the ability to perform aggressive driving when relying on a~\SI{30}{\minute} training dataset.
For increased resilience to sensor failure,~\citet{Tremblay2021} have proposed a multi-modal learned-dynamics model that leverages the various sensor measurements available for~\acp{UGV}. 
Due to the importance of prediction uncertainty in enabling robust control for~\acp{UGV}~\citep{Brunke2022}, this work focuses on~\ac{BLR}, which provides prediction uncertainty estimations~\citep{McKinnon2019}. 
Our novel slip-based \ac{BLR} model allows us to leverage the minimal requirements of learning-based approaches in terms of system characterization,~\citep{McKinnon2019} as well as the improved accuracy of model-based approaches~\citep{Seegmiller2014}.
In this work, the approach of \citet{McKinnon2019} is used as a comparison point, as it is the closest to our model formulation. 

Although both model-based and learning-based approaches require empirical training data, only a few \textbf{dataset-gathering protocols} have been published.
\citet{Voser2010} have proposed to maintain a steady forward velocity while slowly increasing angular velocity, enabling generation of a quasi-steady-state empirical dataset.
\citet{Wang2015} have proposed a similar approach with a varying commanded curvature radius to empirically identify the relation between angular velocity and~\ac{SSMR} skid.
These approaches only cover a small subset of the vehicle's input space.
One can also find large, multimodal datasets allowing to train and evaluate models for off-road and extreme driving~\citep{Triest2022}.
However, such datasets overrepresent forward motion, are limited to a specific~\ac{UGV} and would require new training data for any new vehicle configuration. 
Manual training data gathering guidelines have been proposed by \citet{Williams2018}, asking the driver to vary his driving style.
However, these remain time-consuming and subject to input space coverage bias. 
We demonstrate that training a motion model with the~\ac{DRIVE} protocol allows increased motion prediction performance and fast training dataset gathering.


\section{Methodology and Theory}
\label{sec:metho}
In this section, we provide details on~\DOUGHNUTCALIB, our automated vehicle characterization and training dataset-gathering protocol.
We then describe our proposed slip-based~\ac{BLR} motion model.
Due to the limited number of~\acp{UGV} accessible to us, we focus on \acp{SSMR}.

\begin{figure}[htbp]
	\centering
    \includegraphics[width=0.49\textwidth]{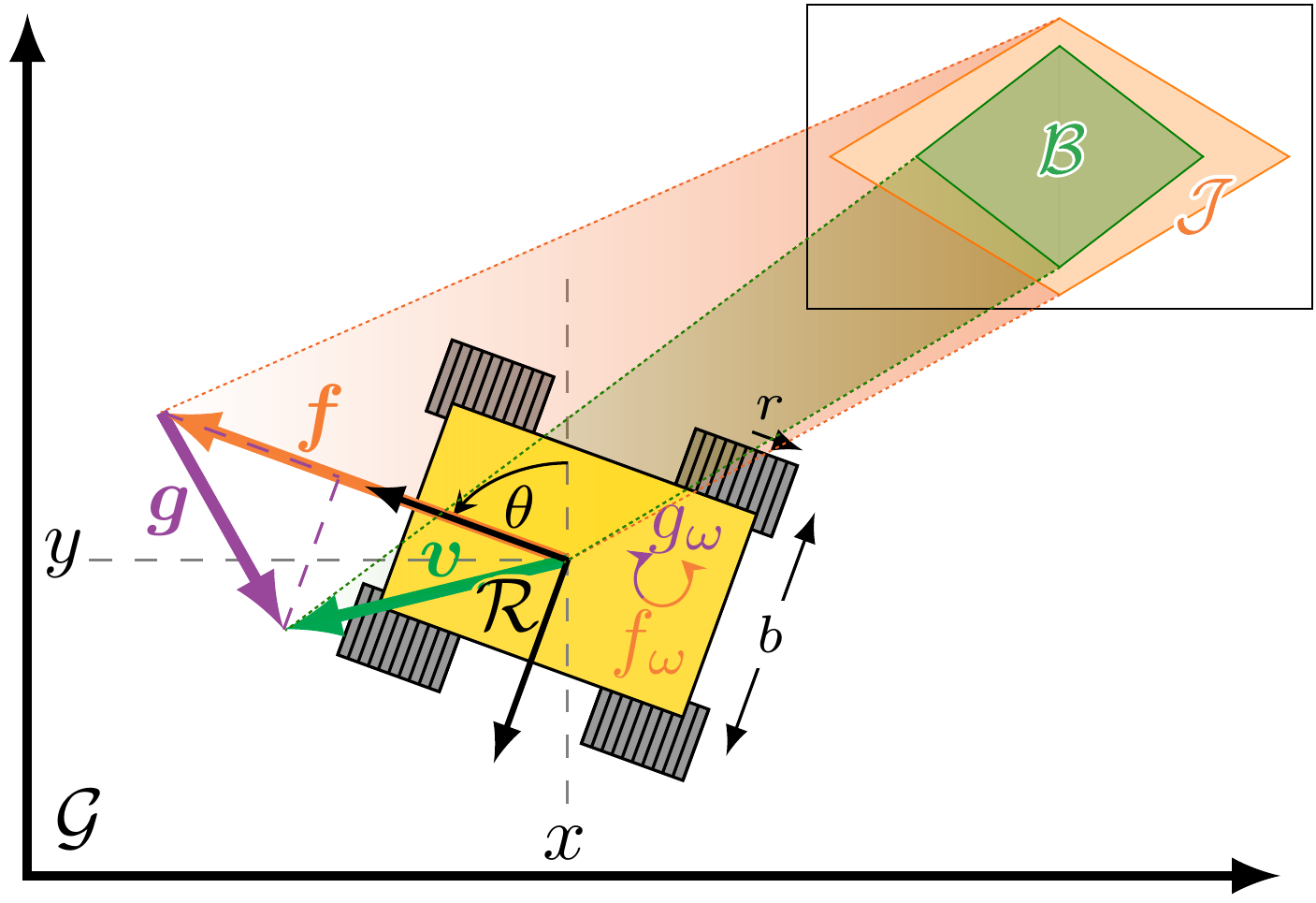} 
     \vspace{-0.3in}
	\caption{Top view drawing of a~\ac{SSMR}.
            In orange is the commanded body velocity~$\CMDBODYVEL$ and the resulting body velocity~$^{\robotf}\bm{v}$ is shown in green.
            The input-space~$\INPUTSPACE$ is shown in orange and the body velocity space~$\BODYVELSPACE$ is shown in green.
            The difference between commanded and resulting body velocity is represented as the slip velocity~$\SLIPBODYVEL$ in purple.
            All represented velocities have an angular component~$(\cdot)_\omega$.
            Robot parameters are the wheel radius~$r$ and vehicle width~$b$. 
            }
	\label{fig:robot-model}
\end{figure}

The involved model variables are depicted in~\autoref{fig:robot-model}.
We limit the states of the vehicle to planar motion, such that the robot's state $^{\mapf}\robotstate = [x, y, \theta]^T$ represents the pose of the vehicle in the global coordinate frame~$\mapf$.
The robot's body frame~$\robotf$, has its $x$ and $y$ axis aligned with the vehicle's longitudinal and lateral directions respectively.
For most~\acp{SSMR}, the input vector is defined as~$\INPUTVECTOR = [\omega_l, \omega_r]^T$, representing the left and right wheel angular velocities.
State propagation, predicting the next state~$\robotstate_{t+dt}$ based on the current state~$\robotstate_{t}$ and input~$\INPUTVECTOR$ is computed as follows: 
\begin{equation}
	^{\mapf}\robotstate_{t+dt} = \ ^{\mapf}\robotstate_{t} + \ \STATEPROPMAT ^{\robotf}\!\bm{v}_{t} \ dt ,
	\label{eq:state_prediction}
\end{equation}
\begin{equation}
	^{\robotf}\bm{v}_{t} = \!\CMDBODYVEL_t(\INPUTVECTOR_{t}) - \!\SLIPBODYVEL_{t} ,
	\label{eq:body_vel}
\end{equation}
where $^{\robotf}\bm{v}_t$ is the vehicle's translational and rotational body velocity, oriented in the vehicle's inertial frame~$\robotf$, and $\STATEPROPMAT$ is a transformation matrix producing a rotation of the robot's angle in the world frame~$^{\mapf}\theta$.
The vehicle's body velocity~$^{\robotf}\bm{v}_{t}\in\mathbb{R}^3$ is modeled as the commanded velocity~$\CMDBODYVEL_t(\INPUTVECTOR_{t})\in\mathbb{R}^3$, from which we subtract the slip velocity~$\SLIPBODYVEL_{t} \in \mathbb{R}^3$. 
The delay between~\ac{UGV} commands is represented by~$dt$.
The diamonds in the top-right inset of \autoref{fig:robot-model} represent conceptually the set of possible commanded~\ac{UGV} body velocities~$\INPUTSPACE$ in orange and the set of actually possible body velocities~$\BODYVELSPACE$.
Thus, to characterize~\ac{UGV} motion, we require two things: a protocol to gather empirical data and a model to learn vehicle slip.
\vspace{-2mm}

Our~\DOUGHNUTCALIB protocol, described in~\autoref{sec:protocol}, automates the task of gathering a complete training dataset~\TRAINDATA$= \left\{\INPUTVECTOR, \tilde{\INPUTVECTOR},  \bm{X}_x, \OBSERVEDSLIP_x, \bm{X}_y, \OBSERVEDSLIP_y, \bm{X}_\omega, \OBSERVEDSLIP_\omega\right\}$, where we concatenate~$n$ observations of slip input~$\bm{X} \in \mathbb{R}^{n \times k}$ and slip velocities~$\OBSERVEDSLIP \in \mathbb{R}^{n}$ for each vehicle dimension.
The number of slip inputs per dimension is defined as~$k$.
This data is also used to train a powertrain model for each side of the~\ac{SSMR} by minimizing the error between commanded~$\tilde{\INPUTVECTOR}$ and measured~$\INPUTVECTOR$ vehicle input through the powertrain parameters described in~\autoref{sec:powertrain}.
We then use~\ac{BLR} to train our slip-based learning model, described in~\autoref{sec:slip_blr}. 
\subsection{\acf{DRIVE}}
\label{sec:protocol}
To learn any~\ac{UGV} slip model, we require a training dataset~\TRAINDATA. 
Details on the composition of this dataset are given throughout~\autoref{sec:powertrain_slip_models}. 
It has been shown that numerous factors impact~\ac{UGV} dynamics, such as vehicle orientation~\citep{Seegmiller2016}, tire saturation~\citep{Djeumou2023} and ground properties~\citep{Baril2020}. 
However, coverage of the entire spectrum of dynamic features would require a large-scale training dataset, and extracting insights from this dataset would be plagued by the curse of dimensionality~\citep{Bellman1966}.
We simplify our problem by enabling stimulation of the entire input space~$\INPUTSPACE$, represented as the orange diamond in~\autoref{fig:robot-model}.
We rely on random, uniform sampling of~\ac{UGV} inputs to ensure broad input space coverage and to trigger dynamic, transitory behavior.
Additionally, it was previously shown that a random search on multidimensional space is faster and offers performance similar to that of a grid search~\citep{Bergstra12}. 
For models accounting for more features, an investigation of more targeted sampling approaches should be conducted.
The system requirements to perform our protocol are as follows:
(i) a sub-servo system mapping body-level commands to wheel commands;
(ii) vehicle acceleration limits to reduce strain on vehicle components;
(iii) an accurate localization system, estimating the robot position and velocity in a global frame; 
(iv) a safety operator to prevent the vehicle from leaving a predefined safe perimeter during the protocol.
Lastly, we release our protocol as an open-source package to facilitate replicability.

\begin{figure}[htbp]
\vspace{0.1in}
	\centering
	\includegraphics[width=0.49\textwidth]{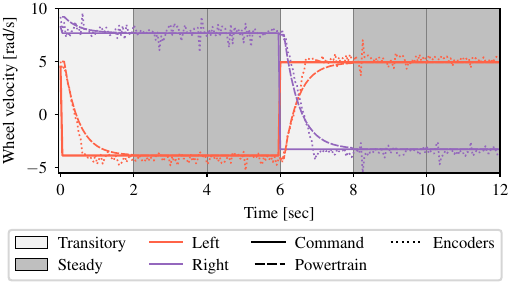}
  \vspace{-0.3in}
	\caption{
            Commanded, encoder-measured and modeled wheel velocities for both sides of a~\ac{SSMR} during two~\DOUGHNUTCALIB training intervals. 
            The powertrain model is described in~\autoref{sec:powertrain}.
            Each training step consists of one transient-state window (in light gray) and two steady-state windows (in dark gray).
            Commands and measurements on the x-axis are acquired at a rate of~\SI{20}{\hertz}.
            }
	\label{fig:transitory_steady_state}
\end{figure}

The first step of~\DOUGHNUTCALIB is to send high body longitudinal velocity commands to the platform in both directions to determine the wheel velocity limits $\omega_{min}$ and $\omega_{max}$. 
This way, we define the~\ac{UGV}'s true input-space $\INPUTSPACE$ as combinations of left and right commanded wheel velocities, such that $\omega_{min} \leq \omega_l, \omega_r \leq \omega_{max}$ for \acp{SSMR}. 
This input-space~$\INPUTSPACE$ is shown in terms of body velocities as the orange diamond in~\autoref{fig:robot-model}. 
Once the input-space limits are defined, we sample input vectors~$\INPUTVECTOR_s$ from our calibration distribution, defined as a two-dimensional uniform distribution parametrized by vehicle input limits $\INPUTVECTOR_s \sim \mathcal{U}_2 \left(\, \bm{\omega}_{min}, \bm{\omega}_{max} \, \right)$.
Our goal is then to maintain each sampled input for a duration sufficient to gather steady-state motion data.
Thus, we define training windows of~\SI{2}{\second}, the same duration used in the seminal~\ac{UGV} path following work of~\citet{Williams2018}. 
We assume the~\ac{UGV} requires one training window to reach any desired wheel velocity.
Since the majority of~\ac{UGV} motion is quasi-steady state~\citep{Voser2010}, we define a training interval consisting of one transient training window and two steady training windows, lasting a total of~\SI{6}{\second}.
An example of two training intervals is shown in~\autoref{fig:transitory_steady_state}.
We use transient-state windows to train our powertrain model, described in~\autoref{sec:powertrain}. 
We keep both steady-state and transient-state windows for our model training and evaluation to enable both steady-state and transient model training.
This procedure is repeated until satisfactory model accuracy is reached.
\autoref{sec:slip-blr_result} presents an analysis of model accuracy with respect to training dataset driving time, indicating model convergence with~\SI{46}{\sec} of driving data.
If the user's goal is simply to gather a large dataset, the protocol can run until~\ac{UGV} battery depletion.
\vspace{-0.1in}
\subsection{Dynamics-aware slip-based model}
\label{sec:powertrain_slip_models}
\subsubsection{Powertrain model}
\label{sec:powertrain}
We start by defining a powertrain model to reduce the predicted wheel velocity~$\hat{\omega}$ error, the result of which can be seen as both dashed lines in~\autoref{fig:transitory_steady_state}.
We use the same first-order plus dead time
transient response model as used by~\citet{Seegmiller2013}:
\begin{equation}
    \begin{aligned}
        \hat{\omega}_t &= \left(e^\beta\right) \omega_{t_0} + \left(1 - e^\beta\right) \tilde{\omega}_{t - \tau_d} \ , \\
        \beta &= \frac{(t - \tau_d)}{\tau_c}  \ ,
    \end{aligned}
	\label{eq:powertrain}
\end{equation}
where $\hat{\omega}$,~$\tilde{\omega}$ and~$\omega$ are the predicted, commanded and measured wheel velocities, respectively. 
We also define the initial time~$t_0$ and prediction horizon at time~$t$.
Here, the parameters that require characterization are the time constant $\tau_c$ and the time delay $\tau_d$.
One should note that these parameters are not considered symmetrical in our protocol and are trained independently for both sides of~\acp{SSMR}.
Thus, our protocol can identify vehicle powertrain asymmetries.
\subsubsection{Body slip model}
\label{sec:slip_blr}
Next, we define a model enabling the computation of both the commanded body velocity~$\CMDBODYVEL_t$ and resulting slip velocity~$\SLIPBODYVEL_t$ with respect to predicted input~$\tilde{\INPUTVECTOR}_t$.
For~\acp{SSMR}, the commanded body velocity~$\CMDBODYVEL_t(\tilde{\INPUTVECTOR}_{t})$ can be modeled through the ideal differential-drive model~\citep{Mandow2007} as 
\begin{equation}
	\CMDBODYVEL_t(\INPUTVECTOR_{t}) = \begin{bmatrix}
		f_x \\
		f_y \\
		f_\omega \\
	\end{bmatrix} = 
	r \begin{bmatrix}
		\frac{1}{2}, \frac{1}{2} \\
		0, 0 \\
		-\frac{1}{b}, \frac{1}{b} \\
	\end{bmatrix} \begin{bmatrix}
		\hat{\omega}_{l_t} \\
		\hat{\omega}_{r_t} \\
	\end{bmatrix} ,
	\label{eq:cmd}
\end{equation}
where $r$ and $b$ are the~\ac{SSMR}'s wheel or track sprocket radius and vehicle width, respectively, as shown in~\autoref{fig:robot-model}.
We use the estimated wheel velocities through~\autoref{eq:powertrain} as the input vector~$\tilde{\INPUTVECTOR}_{t}$.
We consider slip in each dimension of the vehicle separately $\SLIPBODYVEL_t = [g_x, g_y, g_\omega]^T$, with the form
\begin{equation}
	g_t = \blrweights^T \ \! \blrinputs_t + \eta ,
	\label{eq:blr_slip}
\end{equation}
where~$\blrweights \in \mathbb{R}^{k}$ are the weights associated with each slip input and~$\eta \sim \mathcal{N}(0, \sigma^2)$.
We draw inspiration from off-road vehicle dynamics work in the literature to define dynamics-aware basis functions for vehicle slip~\citep{Seegmiller2014}.
As shown by~\citet{Seegmiller2016}, the following set of basis functions to estimate vehicle slip shows similar performance as fully dynamic models in off-road terrain.
Firstly, for longitudinal slip~$\SLIPBODYVEL_x$, we use the vehicle's rolling resistance, proportional to commanded body longitudinal velocity~$\blrinputs_x = f_x$.
Secondly, for lateral slip~$\SLIPBODYVEL_y$, we use centrifugal force~$\blrinputs_y = \CENTRIFUGAL = (f_x f_\omega)$, proportional to commanded longitudinal and angular velocities.
Thirdly, for angular slip~$\SLIPBODYVEL_\omega$, we use three distinct slip learning inputs $\blrinputs_\omega = [\CENTRIFUGAL, f_x, f_\omega]$.
The first angular slip input is the vehicle's centrifugal force~$\CENTRIFUGAL$.
We then add~\ac{UGV} asymmetry, which can be caused by manufacturing imperfections and mechanical wear, causing an angular velocity error proportional to the commanded longitudinal velocity~$f_x$. 
Finally, we account for the vehicle's skid, leading to an error between the commanded angular velocity and the actual angular velocity~$f_\omega$.
It should be noted that the vehicle gravity-dependent parameters, used by~\citet{Seegmiller2014}, are missing in this work. 
The reason is that we simplify our calibration protocol to be executed on planar terrain.
The remainder of this section describes how we learn slip for a single dimension, but the process is the same for all dimensions of slip. 

We use~\acf{BLR} to estimate the values for~$\blrweights$ and~$\sigma^2$.
For a more in-depth explanation of~\ac{BLR}, refer to the book written by~\citet{Murphy2012}.
It can be shown that the posterior for learned parameters $p(\blrweights, \sigma^2 | \mathcal{D}_{\text{d}})$ is distributed according to a Normal Inverse Gamma distribution $\text{NIG}(\blrweights, \sigma^2 | \blrweights, \bm{K}, a, b)$,
where 
\begin{equation}
	\begin{aligned}
		\blrweights &= \bm{K}\left(\bm{K}_0^{-1} \blrweights_0 + {\bm{X}}^T{\OBSERVEDSLIP}\right), \\
		\bm{K} &= (\bm{K}_0^{-1} + {\bm{X}}^T {\bm{X}})^{-1}, \\
		a &= a_0 + \frac{n}{2},  \\
		b &= b_0 + \frac{1}{2}\left(\blrweights_0^T \bm{K}_0^{-1} \blrweights_0 + {\OBSERVEDSLIP}^T{\OBSERVEDSLIP} - \blrweights^T \bm{K}^{-1}\blrweights\right) ,
	\end{aligned}
	\label{eq:posterior_BLR_params}
\end{equation}
where the estimated covariance of the distribution is represented by~$\bm{K} \in \mathbb{R}^{k \times k}$.
Priors for all parameters are defined by the $(\cdot{})_0$ subscript.
We define $\mathcal{D_\text{d}}$ = $\left\{\bm{X}, \OBSERVEDSLIP\right\}$ as a training dataset consisting of vectors of $n$ concatenated observed values for slip inputs~${\bm{X}}$ and observed slip velocities~$\OBSERVEDSLIP$ for a specific dimension.
The posterior equations can be used to train the \ac{BLR} slip model for each dimension based on a training dataset~$\mathcal{D}_{\text{d}}$.
Once the model is trained, we can predict vehicle slip based on $m$ test inputs~$\tilde{\bm{X}} \in \mathbb{R}^{m \times k}$:
\begin{equation}
    p\left(\hat{\OBSERVEDSLIP} | \tilde{\bm{X}}, \mathcal{D}_\text{d}\right) = \mathcal{T}\left({\OBSERVEDSLIP}|{\bm{X}}\blrweights, \frac{b}{a}\left(\bm{I}_m + {\bm{X}}\bm{K}{\bm{X}}^T\right), 2a\right) ,
\label{eq:prediction}
\end{equation}
where~$\mathcal{T}$ is a Student's t-distribution and~$\hat{\OBSERVEDSLIP}$ represents a vector of~$m$ concatenated predicted slip velocities for a specific direction.
We use an uninformative prior to ensure that our protocol requires as little expertise as possible to execute. 
This consists of setting $a_0 = b_0 = 0$, $\blrweights_0 = \bm{0}$ and $\bm{K}_0 = \phi(\bm{X}^T\bm{X})^{-1}$ for any positive value $\phi$.
This allows our slip-based \ac{BLR} model to be initialized with little knowledge of the~\ac{UGV} except for wheel radius~$r$ and vehicle width~$b$.


\section{Results}
\label{sec:results}
In this section, we evaluate the improvement of motion prediction accuracy when training models with the~\DOUGHNUTCALIB protocol. 
We also demonstrate that for off-road navigation of \acp{SSMR}, learning vehicle slip based on dynamics-aware basis functions is more accurate than learning on vehicle acceleration.
Finally, we analyze the number of training data required to reach convergence with our model. 

\subsection{Experimental Setup}
\label{sec:exp_setup}
\begin{figure} [htpb]
	\centering
	\begin{minipage}{0.48\textwidth}
		\includegraphics[width=\textwidth]{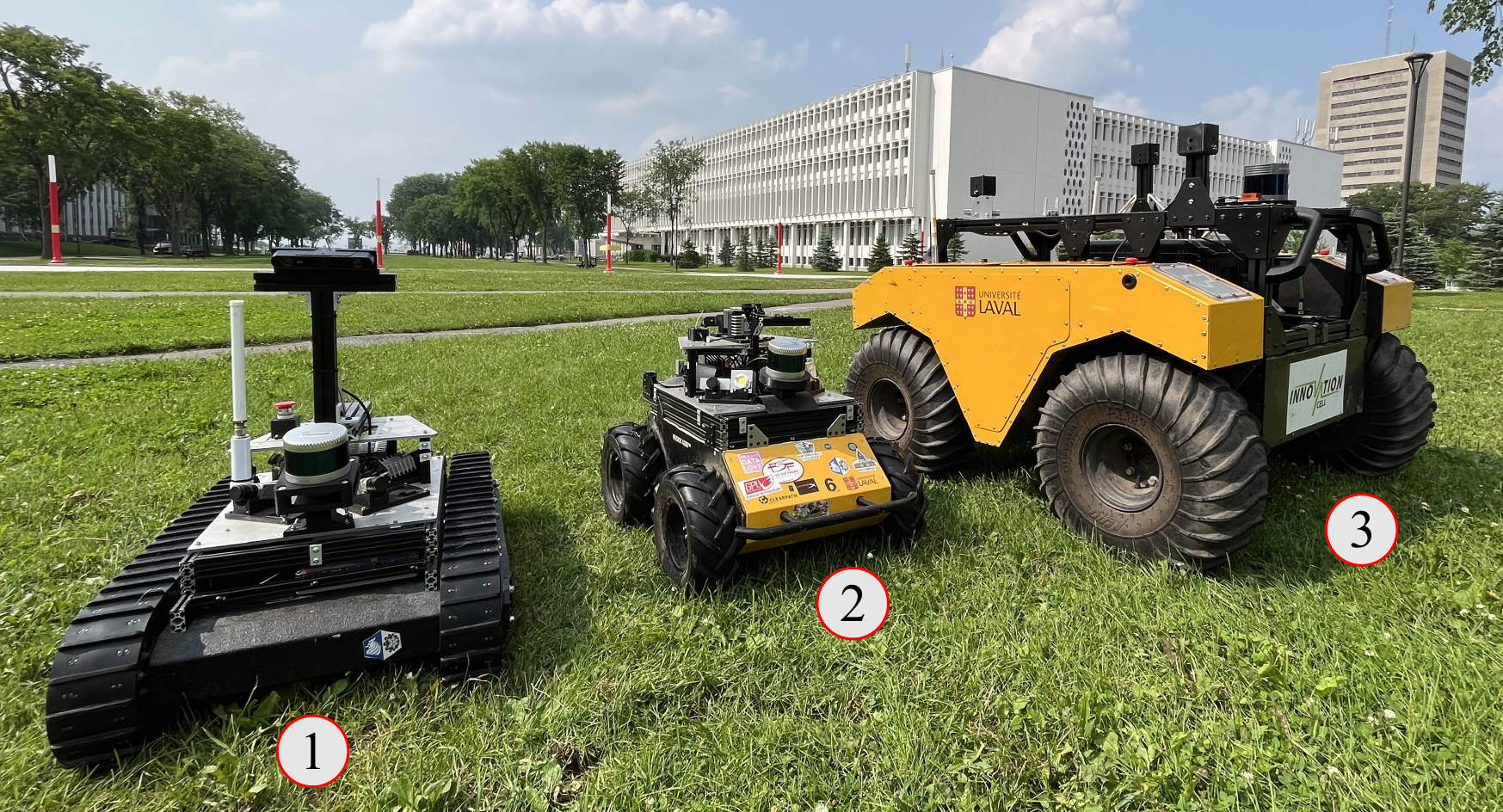}
	\end{minipage}

	\caption{Three different commercial platforms that were used for the experimental work:
	a \emph{Superdroid} HD2 (1), a \emph{Clearpath Robotics} Husky (2), and a \emph{Clearpath Robotics} Warthog mounted
    on wheels (3). 
    The platforms weigh~\SI{80}{\kilo\gram},~\SI{75}{\kilo\gram} and~\SI{470}{\kilo\gram}, respectively.} 
	\label{fig:test_platforms}
\end{figure}

We have conducted experiments using three distinct~\ac{UGV} platforms, as shown in~\autoref{fig:test_platforms}. 
First, we evaluated our protocol and model on a \emph{Clearpath Robotics} Warthog on wheels, weighing~\SI{470}{\kg}, on gravel-covered terrain and an ice rink.
The ice rink was leveled and recently resurfaced, leading to extreme vehicle slip.
Next, we tested our protocol and model on smaller platforms, namely a wheeled \emph{Clearpath Robotics} Husky, weighing~\SI{75}{\kg}, and a tracked \emph{Superdroid} HD2, weighing~\SI{80}{\kg}, both on indoor tile and snow-covered terrain.
The Warthog has a top speed of~\SI{5}{\meter / \second}, which is around five times that of the HD2 at~\SI{1.2}{\meter / \second} and of the Husky at~\SI{1}{\meter / \second}.      
These platforms and terrains were selected to maximize the difference in properties between experiments.
Localization ground truth is estimated through point-cloud registration with the~\ac{ICP} algorithm to use a common, centimeter-accurate ground truth~\citep{Pomerleau2013} across all indoor and outdoor experiments.
The localization system for the Husky and~\MARMOTTE~robots is described in~\citep{Ebadi2024} and for the Warthog in~\citep{Baril2022}.
For every experiment, the recorded data was split into two halves, the training dataset and the evaluation dataset, to enable extensive model evaluation. 
Our experimental dataset totals over~\SI{7}{\kilo\meter} and~\SI{1.8}{\hour} of driving data across all platforms and terrain types. 

\vspace{-2mm}
\subsection{Protocol performance analysis}
\label{sec:protocol_performance}
First, we define a function to evaluate model prediction performance.
While learned models are trained on single-step vehicle slip or acceleration, our goal is to use them to predict vehicle motion over a specific horizon.
We train our model on single-step slip velocities to simplify the learning problem.
\citet{Williams2018} showed that this simplification allows sufficient prediction performances for high-speed~\ac{UGV} path following.
Thus, we use the~\ac{MRMSE}~$\MRSME$ to evaluate prediction errors~\citep{McKinnon2019}, with our localization as ground truth:
%
\begin{equation} 
	\begin{aligned}	
	\MRSME &= \frac{1}{\WINSIZE} \sum_{j=1}^{\WINSIZE} \sqrt{(\MEASSTATE_j - \PREDSTATE_j)^T \bm \Sigma (\MEASSTATE_j - \PREDSTATE_j)}  \ ,
	\end{aligned}
	\label{eq:MRSME}
\end{equation}
where $\WINSIZE$ is the prediction window size. 
We define the measured state as $\MEASSTATE$ and the model-predicted state as $\PREDSTATE$.
All robots are commanded at~\SI{20}{\hertz} and the prediction window length is set at~\SI{2}{\second} reflecting the established path following work of~\citet{Williams2018}.
We divide this error into translational \ac{MRMSE}~$\MRSME_T$, for which~$\bm \Sigma = \text{diag}(1, 1, 0)$ and rotational~\ac{MRMSE}~$\MRSME_R$, for which~$\bm \Sigma = \text{diag}(0, 0, 1)$.

\begin{figure}[htbp]
	\centering
	\includegraphics[width=0.49\textwidth]{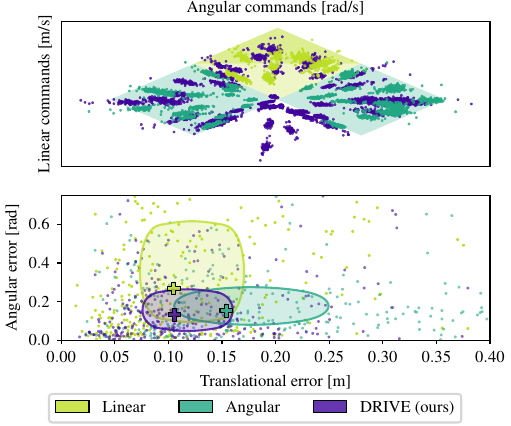} 
    \vspace{-0.3in}
	\caption{
            Data-gathering protocol performance for the~\MARMOTTE on snow experiment.
            The top subplot illustrates the three data-gathering methods compared in this work.
            In yellow, we have the linear-focused method.
            In teal, we have the angular-focused method.
            In blue-violet is our~\DOUGHNUTCALIB approach.
            The crosses and regions on the bottom subplot show the medians and interquartile ranges for translational and angular prediction errors.}
	\label{fig:prediction_improvement_step} 
\end{figure}

We compute~\ac{MRMSE} for three distinct training datasets gathering approaches. 
\autoref{tab:drive_improvement} shows the median prediction improvement for all robots and terrains.
Translational and rotational prediction improvements are computed by comparing our~\DOUGHNUTCALIB approach with the angular-focused~\citep{Wang2015} and linear-focused~\citep{Williams2018} approaches, respectively. 
The Warthog on ice experiment stands out due to the high prediction error, which is discussed in~\autoref{sec:slip-blr_result}.
Without considering this experiment,~\DOUGHNUTCALIB offers a mean translation improvement of~\SI{31.8}{\%} over the angular-focused approach and a mean rotation improvement of~\SI{43.6}{\%} over the linear-focused approach. 
\autoref{fig:prediction_improvement_step} focuses on the HD2 on snow experiment, showing all three gathering approaches and the resulting prediction error.
It can be seen that the~\DOUGHNUTCALIB protocol significantly outperforms other training data-gathering methods, both for translation and rotation.
\begin{table}[!ht]
	\caption{
		Median prediction improvement for all robots and terrains.
	}
	\label{tab:drive_improvement}
    \centering
	\begin{tabularx}{0.48\textwidth}{l*{6}{>{\centering\arraybackslash}X}}
			\toprule
			Prediction & \multicolumn{2}{>{\hsize=\dimexpr2\hsize+2\tabcolsep+\arrayrulewidth\relax}c}{Husky}
			& \multicolumn{2}{>{\hsize=\dimexpr2\hsize+2\tabcolsep+\arrayrulewidth\relax}c}{HD2}
			& \multicolumn{2}{>{\hsize=\dimexpr2\hsize+2\tabcolsep+\arrayrulewidth\relax}c}{Warthog} \\
			improvement & Tile & Snow & Tile & Snow & Gravel & Ice  \\ \midrule
			Translation (\%) & 35 & 10 & 33 & 31 & 50 & 11 \\
			Rotation (\%) & 18 & 61 & 27 & 51 & 61 & 7 \\
			\bottomrule
		\end{tabularx}
\end{table}
\vspace{-5mm}
\vspace{-2mm}
\subsection{Slip-based learning predictive performance}
\label{sec:slip-blr_result}
\begin{figure}[htbp]
	\centering
 	\includegraphics[width=0.48\textwidth]{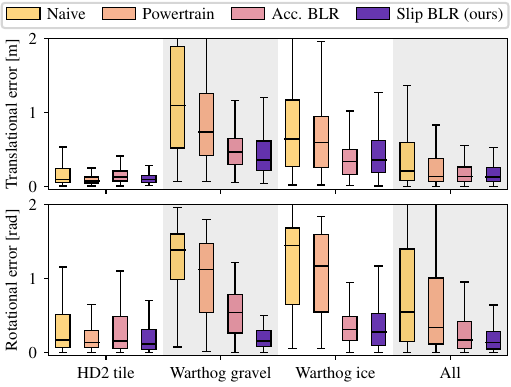} 
	\caption{
            Translational and rotational prediction errors for all models studied in this work.
            In yellow is the manufactured-defined naive model, in orange is the powertrain-aware model described in~\autoref{sec:powertrain}, in red is the acceleration-based~\ac{BLR} model and in purple is our slip-based~\ac{BLR} model. 
    }
	\label{fig:prediction_benchmark}
\end{figure}

In~\autoref{sec:powertrain_slip_models}, we propose a novel slip-based~\ac{BLR} model to predict \ac{UGV} motion.
\autoref{fig:prediction_benchmark} shows this model's performance for both translational and rotational prediction, compared to the model proposed by~\citet{McKinnon2019}, which performs~\ac{BLR} on~\ac{UGV} actuator dynamics.
We present three distinct datasets, namely~\MARMOTTE on tile along with the wheeled Warthog on gravel and ice. 
The rightmost results combine the prediction errors for all experiments conducted in this work. 
We also show the performance of the model provided by manufacturers (i.e., Naive) and the improvement achieved through powertrain modeling.

When accounting for all datasets, we observe a~\SI{34}{\%} decrease in translation prediction error median and a~\SI{38}{\%} decrease in rotation prediction error median when comparing the naive model with the powertrain-aware model.
Also, our slip-based~\ac{BLR} approach leads to a~\SI{22}{\%} decrease in rotation prediction error median and a~\SI{6}{\%} decrease in translation prediction error median when compared to acceleration-based~\ac{BLR}. 
The Warthog in gravel shows the largest improvement between our slip~\ac{BLR} and acceleration~\ac{BLR}, with~\SI{71}{\%} in rotation error median and~\SI{23}{\%} in translation error median. 
In contrast, the~\MARMOTTE on tile experiment shows a performance decrease for acceleration~\ac{BLR} and similar performance for slip~\ac{BLR} when compared to the powertrain model. 
Indeed, the indoor tile ground already had a low prediction error for the powertrain-aware model.
Lastly, the ice rink experiment shows a similar performance between slip and acceleration~\ac{BLR}.
This experiment corresponds to extreme slip, similar to a~\ac{UGV} driving over black ice for an extended duration. 
This result shows the limit of our slip-based~\ac{BLR} model which still performs similarly or better than other models.
In this case, dynamic modeling could improve performance. 
Overall, we conclude that slip-based~\ac{BLR} offers improved performances for rotation prediction and similar performance in translation prediction over acceleration-based~\ac{BLR}, especially for driving at higher velocities on off-road terrains. 
For~\acp{SSMR} in particular, rotation motion is the highest source of error due to the complexity of wheel-terrain skidding interactions~\citep{Baril2020}, justifying the significance of our model. 

\begin{figure}[htbp]
	\centering
	\includegraphics[width=0.48\textwidth]{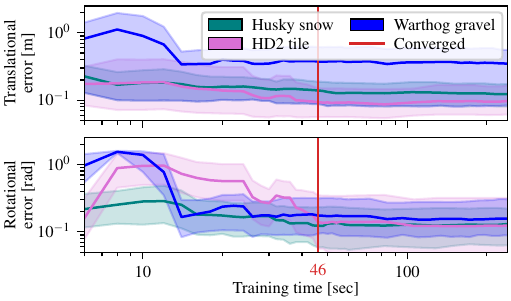} 
 \vspace{-0.25in}
	\caption{
            The relation between training driving time and our slip-based~\ac{BLR} model prediction performance, for translation and rotation. 
            Three datasets are shown, namely the Husky on snow in teal, the HD2 on tile in pink and the wheeled Warthog on gravel, in blue.
            We highlight at~\SI{46}{\second} the converged value with the red line, for which our model converges for all~\acp{UGV} tested.
            For all subplots, both axis are in log scale.
    }
	\label{fig:training_time}
\end{figure}

Moreover, generating the training data is time and energy-consuming, which leads us to look for a trade-off between calibration duration and model prediction accuracy.
Thus, we evaluated the relationship between training driving time and prediction accuracy.
The results are shown in~\autoref{fig:training_time}.
Three distinct experiments are presented, notably Husky on snow, HD2 on tile and Warthog on gravel.
No other experiment is shown, to avoid cluttering, but similar results were observed. 
As specified in~\autoref{sec:slip_blr}, an uninformative prior is used for every platform, explaining the initially high errors.

As shown by the red vertical line in~\autoref{fig:training_time}, the prediction accuracy stabilizes after a maximum of~\SI{46}{\second} of driving time for all experiments. 
To compute this time, we evaluated the maximum driving time for which the gradient of translational and rotational prediction error with respect to training driving time was under~\SI{0.01}{\meter / \second} and~\SI{0.01}{\radian / \second} for all shown experiments, indicating all models have converged.
Thus, users of the~\DOUGHNUTCALIB protocol~\acp{SSMR} can expect that the slip-based~\ac{BLR} motion model has converged after~\SI{46}{\second} of training data. This is almost four times shorter than~\SI{180}{\second}, which is shortest training time reported in the literature~\citep{Djeumou2023}. 

\section{Conclusion}
\label{sec:conclusion}
In this paper, we propose~\emph{\acf{DRIVE}}, an automated vehicle characterization and training data generation protocol.
We also propose a novel~\ac{UGV} prediction model called slip-based~\ac{BLR}.
We show that training our model with our protocol offers improved prediction performances when comparing common training approaches and similar learning-based models.
We also show that with our protocol, model convergence is reached with four times less driving time than the shortest similar protocol. 
Future work would include generalizing our protocol to any vehicle geometry (e.g., Ackermann steering) and adapting our model formulation for complete dynamic models for extreme slip situations such as driving on surfaced ice.
Various input sampling strategies should be investigated in conjunction with dynamic models to further minimize training dataset driving time.

\addtolength{\textheight}{-0.42cm}   

\printbibliography

\end{document}